\documentclass[runningheads]{llncs}

\usepackage{cite}
\usepackage{amsmath,amssymb,amsfonts}
\usepackage{algorithm}
\usepackage{algpseudocode}
\usepackage{graphicx}
\usepackage{textcomp}
\usepackage{xcolor}
\usepackage{paralist}
\usepackage{array}
\usepackage{tabularx}
\usepackage[inline]{enumitem}
\usepackage{booktabs}
\usepackage[table]{xcolor}
\usepackage{multirow}
\usepackage{verbatim}
\usepackage{makecell}
\usepackage{caption}
\usepackage{subcaption}
\usepackage{tikz}

\def\BibTeX{{\rm B\kern-.05em{\sc i\kern-.025em b}\kern-.08em
    T\kern-.1667em\lower.7ex\hbox{E}\kern-.125emX}}

\newcommand\copyrighttext{%
	\footnotesize This is the preprint accepted for publication in the 22nd EAI International Conference on Mobile and Ubiquitous Systems: Computing, Networking and Services (MobiQuitous), Shanghai, China, 7-9 November 2025. This version is released under a CC-BY license according to the requirements of the Horizon Europe programme that has provided funding for this work.}
\newcommand\copyrightnotice{%
	\begin{tikzpicture}[remember picture,overlay]
		\node[anchor=north,yshift=-10pt] at (current page.north) {\fbox{\parbox{\dimexpr\textwidth-\fboxsep-\fboxrule\relax}{\copyrighttext}}};
	\end{tikzpicture}%
}

\begin{document}

\title{
Using Machine Learning to Take Stay-or-Go Decisions in Data-driven Drone Missions
}

\titlerunning{Using Machine Learning to Take Stay-or-Go Decisions}

\author{Giorgos Polychronis \and
Foivos Pournaropoulos \and \\
Christos D. Antonopoulos\and
Spyros Lalis}
\institute{University of Thessaly, Volos, Greece \\
\email{\{gpolychronis, spournar, cda, lalis\}@uth.gr}}

\authorrunning{Giorgos Polychronis et al.}

\maketitle

\copyrightnotice

\begin{abstract}
Drones are becoming indispensable in many application domains. 
In data-driven missions, besides sensing, the drone must process the collected data at runtime to decide whether additional action must be taken on the spot, before moving to the next point of interest. 
If processing does not reveal an event or situation that requires such an action, the drone  
has waited in vain instead of moving to the next point.  
If, however, the drone starts moving to the next point 
and it turns out that a follow-up action is needed at the previous point, it must spend time to fly-back. To take this decision, we propose different machine-learning methods based on branch prediction and reinforcement learning. We evaluate these methods for a wide range of scenarios where the probability of event  
occurrence changes with time. Our results show that  
the proposed methods consistently outperform the regression-based method  
proposed in the literature and can significantly improve the worst-case mission time by up to $4.1x$. 
Also, the achieved median mission time is very close, merely up to $2.7\%$ higher, to that of a method with perfect knowledge of the current underlying event probability at each point of interest.     
\end{abstract}


\section{Introduction}

Drones, multicopters in particular, have  
become  
popular across a wide range of civilian applications, 
because they are easy to deploy, they can fly/hover in a very controllable way, and can be equipped with various sensors. 
In several cases, the missions 
are data-driven, 
i.e., the drone may need to 
perform further action(s) depending on the data collected via its onboard sensors. 
For example, in a smart agriculture scenario, if pest is detected at a specific location in the field, the drone can  
directly spray 
that location with pesticide. 
In search and rescue missions, 
if a person is detected with some confidence, the drone may  
repeat the sensing from a lower altitude 
and 
deliver a first-aid kit before help arrives.  
In the case of firefighting, 
the same drone could be used to both detect and control  
the fire at its early stage.  
However, the resource- and power-constrained embedded computing platforms of such drones may take a long time to 
process the sensor data to detect events or situations that require further handling. If this processing must be done often, 
the overall mission time can increase significantly.  

One way to reduce the mission time is to 
accelerate data processing by leveraging 
external powerful computational resources (e.g., cloud or edge servers) to offload processing~\cite{messous2020edge},\cite{icfecKasidakis}, \cite{polychronis}. 
A complementary approach, proposed in% 
~\cite{staygo}, is to exploit the fact that a follow-up action may not always be needed at every point of interest. 
Namely, the drone can 
proceed with its mission and  
go to the next point of interest, right after the sensing task is completed, to overlap the computation time with the flight time. 
If, however, 
it turns out that  
an action is needed at the previous point of interest,  
extra time  
is  
spent for the 
drone to fly back. 
The  
alternative is for the drone  
to stay at the point of interest and wait for the computation to finish. 
But if no follow-up action is required, 
the drone will have wasted time waiting in vain. 

In this paper, we focus on the problem of learning how to take  
good stay-or-go decisions. 
More specifically, 
the main contributions  
are: \begin{inparaenum}[(i)] 
\item We propose a perceptron-based approach to tackle the decision 
problem. 
\item In addition, we tackle the problem using  
reinforcement learning.
\item We evaluate both approaches via extensive simulation experiments  
for a wide range of scenarios where the probability of action at each point of interest 
changes with time. 
\item Our results show that both  
approaches can achieve good results in dynamic environments, clearly outperforming the regression-based method proposed in~\cite{staygo} while performing close to a method that takes decisions based on perfect knowledge of the underlying probabilities.
\end{inparaenum}

The structure of the paper is the following. Section~\ref{sec:rel} gives an overview of related work. Section~\ref{sec:prob} presents the system model and  
the decision problem. 
Section~\ref{sec:exec} provides the logic for controlling the drone to perform the mission at hand, independently of the method that is used to take the stay-or-go decision at each point of interest. 
Section~\ref{sec:brachpred} and Section~\ref{sec:ml}  
describe the  
methods for taking such decisions based on  
branch prediction 
and reinforcement learning, respectively. 
Section~\ref{sec:eval} presents the evaluation of the 
proposed decisions methods. 
Finally, Section~\ref{sec:concl} concludes the paper.
\section{Related Work}
\label{sec:rel}

\textbf{Reducing mission time by offloading.}
A number of studies aim to reduce the completion time of resource-intensive computations on autonomous drones by offloading them to the cloud or an edge infrastructure. 
For example, 
the authors of~\cite{messous2020edge} focus 
on drone-based navigation and mapping in unknown areas, where the drone dynamically decides when to offload processing.
\cite{chen2020intelligent} proposes an algorithm that enables multiple drones to dynamically select edge servers for task offloading, taking into account factors such as channel quality and predicted trajectory. 
In~\cite{icfecKasidakis}, the drone uses a heuristic to choose at runtime between executing the computation locally or offloading it to an edge server. The choice relies on prior knowledge of each server’s end-to-end response time.
\cite{polychronis} 
investigates the combined mission planning and offloading for the case where multiple drones  
executing different missions experience uncertainty in flying times, while offloading computations to nearby edge servers to reduce the time waiting for the results. 
Our approach complements these studies by addressing mission time optimization through informed decision-making about whether the drone should wait for computation results or proceed to the next waypoint. 
The authors of \cite{bandarupalli2023vega}  
investigate a drone system that starts with high-altitude surveillance to scan large areas. If an object of interest is detected, the drone descends to conduct a more accurate inspection. Their work primarily explores the trade-offs among detection delay, area coverage, and sensing quality. In contrast, our work focuses on the mission time minimization by enabling the drone to proceed to the next point of interest before the current detection results are available.

\textbf{Reinforcement learning.}
ML approaches have  
been used  
to perform  
various optimizations in drone-based systems~\cite{abubakar2023survey, bithas2019survey}. 
In particular, reinforcement learning (RL) is an emerging technique  
used  
in various  
scenarios, offering increased adaptivity and flexibility in dynamic environments compared to classic ML methods. 
For example, several works investigate path planning, navigation, and control~\cite{azar2021drone, li2022path, dabiri2023uav}, as well as computation offloading and resource allocation~\cite{qu2021dronecoconet, NABI2024101342, peng2020multi}. The major differentiation of our work is that we focus on a different problem, where RL is used to learn the optimal decision (stay or go) at the points of interest of data-driven missions to reduce the total mission time, but the correctness of such decisions depends on unknown probabilities that may also change over time. 
In our work, we use reinforcement learning to take the stay-or-go decisions. More precisely, we use the DQN~\cite{mnih2013playing}, 
A2C~\cite{mnih2016asynchronous}
and PPO~\cite{schulman2017proximal} 
algorithms,  
which have good performance in other problems. An extensive comparison  
between these algorithms is provided for the BreakOut Atari game environment in~\cite{de2024comparative}. However, we evaluate these  
algorithms for a completely different problem.

\textbf{Branch prediction.}
During program execution, branch prediction is used to guess the outcome of a branch, so that  
one can  
apply parallelism in the instruction 
execution of a program and improve 
performance~\cite{mcfarling1993combining}\cite{mittal2019survey}\cite{joseph2021survey}. 
Some works  
use perceptrons to predict the outcome of a branch~\cite{jimenez2001dynamic}\cite{jimenez2002neural} or to estimate the branch confidence~\cite{akkary2004perceptron}. Perceptrons are simple neural networks, that are trained to find correlations between the current branch and other branches, or between recent outcomes and old outcomes of the branches. 
In~\cite{seznec2004gehl} the O-GEHL predictor was introduced. This predictor used multiple tables, while the prediction is done in a perceptron-like summation. 
The TAGE predictor~\cite{seznec2006case}, uses multiple tagged predictors each using different history lengths. At prediction, the predictor with the longest history that match the tag at hand is used. 
The problem we tackle in this work, has similarities to the branch prediction problem, as we also want to predict the outcome of processing to overlap computation time with flight time (in case of a go prediction). To this end, we employ a perceptron-based approach. As an additional reference, we use a simpler, 2-bit branch predictor~\cite{lee1984branch}.

The stay-or-go decision problem has been studied in~\cite{staygo} using a regression method with memory-reset logic to delete potentially misleading historical information. 
In this paper, we tackle the problem using 
different prediction approaches (based on branch prediction and RL), showing that they can significantly outperform the regression method in  
scenarios where the probability for action-taking changes.
\section{Problem Formulation}
\label{sec:prob}

The problem we  
address  
was  
introduced in~\cite{staygo}. For the sake of completion, we also provide a short  
description here.

\textbf{Mission.}  
The drone  
visits the points of interest by following a  
pre-specified path, 
encoded as a sequence of waypoints $[wp_1, wp_2, ..., wp_N]$. 
Waypoints $wp_1$ and  
$wp_N$  
correspond to the points from where the drone takes-off and lands, respectively. 
The  
other waypoints in the path, $wp_i, 2 \leq i \leq N-1$,  
stand for the actual points of interest. At each of these points of interest, the drone must capture the required mission-related data, using its onboard sensors and then process this data.
Let $senseT$ and $procT$ denote the time needed to perform the required sensing and data processing computation at each point of interest. 
As a result of this computation, an event or situation may be detected that requires a follow-up action at the respective point of interest. Note that both the computation to be performed and the type of events that may need further handling, both depend on the application at hand. Without loss of generality, let  
$e_i \in \{0,1\}$ indicate whether a follow-up action is needed at $wp_i$ ($e_i$=1), or not ($e_i$=0).  
Also, let the   
time required to perform this action be $actT$. 

\textbf{Flight Model.} Let $flyT_{i,j}$ be the time  
needed for the drone to fly 
from $wp_i$ to 
$wp_j$ as given in Equation~\ref{eq:flyT} below: 
\begin{equation}
\label{eq:flyT}
 flyT_{i,j} = \begin{cases}
   cruiseT_{i,j} + takeoffT, & \text{if } i=1\\
   cruiseT_{i,j} + landT, & \text{if } j=N\\
   cruiseT_{i,j}, & \text{if } i \neq 1 \land j \neq N
 \end{cases}
\end{equation}
\noindent where, $cruiseT_{i,j}$ is the time for  
flying from $wp_i$ to $wp_j$ at cruising speed, $takeoffT$ is the time for a vertical take-off until the drone reaches the desired mission altitude, and $landT$ is the time for vertical landing. 
If the drone  
decides to  
fly to the next waypoint $wp_{i+1}$ 
without waiting for the computation to complete, and an event is detected, $e_i=1$, for the previous point $wp_i$, it must immediately stop its flight toward $wp_{i+1}$ and return to the previous waypoint $wp_i$ to perform the required action. Let $retT(i,procT)$ capture the respective delay, 
as a function of $procT$ 
which determines the travelled distance 
by the time the computation finishes and the  
event is detected.

\textbf{Mission Time.} Let $d_i \in \{0,1\}$ encode the decision taken by the drone whether to stay at $wp_i$ (waiting for the computation to finish) or to go  
to $wp_{i+1}$ (before the computation finishes). 
Then, Equation~\ref{eq:visitT} gives the time that is needed to perform all the required tasks for each visited point of interest: 

\begin{equation}
\label{eq:visitT}
\begin{split}
    visitT_i = senseT + 
   \begin{cases}
    procT + e_i*actT, & \text{if } d_i=0\\
    e_i*(procT + retT(i,procT) + actT), & \text{if } d_i=1\\
  \end{cases}
\end{split}
\end{equation}

\noindent Note that $visitT_i$ always includes $senseT$.  
In addition, if the drone decides to stay, it includes the processing time $procT$ and the time for taking the additional action $actT$ if 
processing indicates an event 
($e_i=1$). If the drone decides to go and this decision proves to be wrong ($e_i=1$), $visitT_i$ includes $procT$, the time to return to the previous point $retT(i,procT)$ and the time to perform the additional action $actT$. 
Finally, Equation~\ref{eq:missionT} gives the time that is needed to complete the entire mission. 

\begin{equation}
\label{eq:missionT}
\begin{split}
    missionT = \sum_{i=1}^{N-1}{flyT_{i,i+1}} + \sum_{i=2}^{N-1}{visitT_i}
\end{split}
\end{equation}

\noindent This is the total flight time required to  visit all waypoints according to the specified path, 
plus the sum of the time spent for each point of interest to perform the required tasks.

\textbf{Objective.} We assume that the same mission is performed repeatedly over a longer time period. 
The objective is for the drone to exploit the experience from previous missions and \textit{learn} to take a good decision $d_i$ at each point of interest $wp_i, 2 \leq i \leq N-1$, so as to minimize $missionT$. 
The challenge is to achieve this without having any knowledge about the underlying probabilities which determine $e_i$ (whether an additional action needs to be taken at $wp_i$). This becomes even harder if these probabilities 
change with time. Next, we describe the control logic of the drone along with two different approaches for taking the stay-or-go decision at each point of interest. 
\section{Mission Execution}
\label{sec:exec}

At each point of interest, the drone must take a decision, to stay or go. Depending on the actual event generation, this decision may turn out to be correct or wrong. Nevertheless, the drone can be programmed to execute the mission in a generic way, independently of the method that is used to take these decisions. It suffices to handle any possible outcome of each decision as needed.

\begin{algorithm}[!h]
\caption{Drone control logic.}
\label{alg:control}
\begin{algorithmic}
\State $\Call{DecisionMethod.MissionBegin}{wp[]}$
\State Autopilot.Arm\&TakeOff()
\Statex
\For{$i \textbf{ from } 2 \textbf{ to } len(wp)-1$}
    \State Autopilot.WaitToArrive()    
    \State $data \gets $ Sense()
    \State StartComputation($data$)
    \State $d_i \gets \Call{DecisionMethod.Decide}{wp_i,procT}$
    \If{$(d_i = 1)$}  \Comment{do not wait, go to the next point of interest}
        \State Autopilot.GoTo($wp_{i+1}$)
    \EndIf
    \State $e_i \gets $ GetComputationResult()
    \State $\Call{DecisionMethod.Feedback}{wp_i,d_i,e_i}$
    \If{($e_i = 1)$}
        \If{$(d_i = 1)$} \Comment{return back to perform action}
            \State Autopilot.GoTo($wp_i$)
            \State Autopilot.WaitToArrive()    
        \EndIf
        \State PerformAction() \Comment{handle the detected event/situation}
    \EndIf
    \If{$(d_i = 0) \lor (e_i = 1)$}  \Comment{go to the next point of interest}
        \State Autopilot.GoTo($wp_{i+1}$)
    \EndIf
\EndFor
\Statex
\State Autopilot.WaitToArrive()
\State Autopilot.Land\&Disarm()
\State $\Call{DecisionMethod.MissionEnd}{ }$
\end{algorithmic}
\end{algorithm}

Algorithm~\ref{alg:control} gives a high-level description of the mission logic used to control the drone by sending corresponding commands to the autopilot. In a nutshell, the drone takes-off and starts visiting the points of interest according to the specified mission plan. When it arrives at a point of interest, it performs the sensing task and starts the computation to process the data collected. Then, the drone decides whether to stay waiting for the computation to finish, or go to the next point. In the first case, the drone just waits hovering above the point of interest. If the computation generates a detection event then the drone performs the necessary action, before moving on. In the second case, the autopilot is instructed to start moving to the next waypoint and the drone waits until it receives the results of the computation. If no event is generated, the mission proceeds as usual, else the autopilot is instructed to go back to the previous waypoint where the drone performs the required action. 
After visiting the last point of interest, the autopilot is instructed to return and land the drone.

The method used to take the stay-or-go decision at each point of interest is abstracted  
as a separate component (\textsc{DecisionMethod}), which is invoked by the mission program through a structured API.  
Note that  
this component can have rich and potentially persistent internal state, 
e.g., to encode the experience gained from previous mission executions so that this  
can be used to take better decisions in the next mission. More specifically, this state is initialized (but not necessarily reset) via method $\Call{MissionBegin}$ before the mission starts and it is updated via $\Call{MissionEnd}$ after the mission ends. The decision at each point of interest is taken via method $\Call{Decide}$ and feedback regarding the correctness of this decision is communicated back to the component via $\Call{Feedback}$. This abstraction simplifies experimentation with different decision methods without changing the core logic of the mission program.

In the following sections, we discuss the various ML-based decision methods we evaluate in this paper. Each method is packaged as a different implementation of the \textsc{DecisionMethod} component, without changing the basic mission execution logic of the drone.

\section{Perceptron  
Approach}
\label{sec:brachpred}

Motivated by work done on branch-prediction~\cite{jimenez2001dynamic}, 
for each point of interest, we use a 1-layer perceptron 
that exploits the experience gained at nearby points of interest. 

Let  
$\mathcal{N}_i = \{wp_j: dist(wp_i, wp_j) < D\}$  
denote the 
neighbors of $wp_i$ (including $wp_i$ itself), 
where $dist()$ returns the distance between two points, and $D$ is the  
distance threshold for the neighborhood boundary. 
Each $wp_j \in \mathcal{N}_i$ contributes to the predictor for $wp_i$ with each own $H$ most recent outcomes $x_{j,h}, 1 \leq h \leq H$.  
More specifically, if in the $h$th previous mission no event was detected at $wp_j$ then $x_{j,h}=-1$, else $x_{j,h}=1$. 
The output value $y_i$ is calculated using Equation~\ref{eq:perceptron}, as the weighted sum of these input values and a bias value $B$ (a small value that effectively comes into play only in the initial case  where  
there is no previous experience,  
to force  
a default decision). 
The final decision for $wp_i$, is \textit{stay} if $y_i>0$, else the decision is \textit{go}. 

\begin{equation}
\label{eq:perceptron}
\begin{split}
    y_i = B + \sum_{j \in \mathcal{N}_i}{\sum_{h=1}^{H}{x_{j,h}\times q_{j,h} \times w_{j,h}}}
\end{split}
\end{equation}

We use two types of weights to factor-in $x_{j,h}$. 
The first type of weights $q_{j,h}$  
capture the gravity of the outcome $x_{j,h}$ in the prediction (the delay caused by a wrong stay decision can be different than the delay caused by a wrong go decision). More specifically, if $x_{j,h}=-1$, we set $q_{j,h} = procT$, else $q_{j,h} = retT(i, procT)$. 
The second type of weights $w_{j,h}$ capture the correlation between different points of interest. 

At the start of each mission, all $w_{j,h}$ are reset to $1$ and then re-trained using the outcomes of the $H$ previous missions. More specifically, we increase or decrease $w_{j,h}$ by $\eta$ depending on whether $x_{j,h}$ was equal to the most recent outcome $x_{i,1}$ ($\eta$ is the so-called learning rate). The reason we reset each $w_{j,h}$ before re-training, rather than just updating its value from the previous mission, is because the correlation between the neighboring points of interest may change. In such case, we wish to let the value of the weights be affected only by the outcome of the $H$ most recent missions (instead of all missions that have been performed so far). 

This initialization step is performed from within the $\Call{MissionBegin}$ method of the \textsc{DecisionMethod} component. The $\Call{Decide}$ method returns the prediction made for the corresponding point of interest based on the current state of the perceptron, while $\Call{Feedback}$ records the actual outcome at that point. Finally, method $\Call{MissionEnd}$ commits the recorded outcomes of the mission so that they can be used in the initialization phase when starting the next mission.

\section{Reinforcement Learning Approach}
\label{sec:ml}

As an alternative  
machine-learning approach for predicting the 
(stay or go) decision that is more likely to be  
beneficial at each  
point of interest in terms of mission time savings, we use reinforcement learning (RL). 
The design of the RL agent is described in more detail below.

\subsection{Training of RL agent}
 
In the 
first mission, the RL agent 
takes decisions without any prior knowledge. 
When the 
mission is completed, the agent is trained  
based on the recorded detection events at  
each point of interest.  
This procedure is repeated in all  
subsequent missions. Note, however,  
that 
the agent is not trained from scratch 
but it is re-trained based on the most recent experience (from the previous mission). 
This way we exploit continual learning to let the agent adapt more easily to changes. 

\subsection{State space, action space and reward function of RL agent}

We perform a discretization of the 
geolocations  
that correspond to the points of interest visited by the drone, to integers. This   
reduces  
state space complexity,  
which, in turn, significantly improves  
learning efficiency. 
Thus the state 
is an integer $i, 2 \leq i \leq N-1$,
where $i$ 
denotes the corresponding point of interest $wp_i$. 
Note that we exclude 
$wp_1$ and $wp_N$  
as these represent the drone's take-off and landing points (the agent 
takes stay/go decisions only at the points of interest).
The action space includes  
the two possible  
decisions that can be taken at each  
$wp_i$, namely  
to stay ($d_i=0$) or to go ($d_i=1$). 

The reward function is designed to provide feedback to the RL agent based on  
the experienced  
time gains or penalties, 
depending on whether the decision was correct or not. 
The reward $r$ 
is given by Equation~\ref{eq:rl_reward} as a function of the state 
$i$, the presence of an event that must be handled $e_i$, and the agent's decision $d_i$: 

\begin{equation}
\label{eq:rl_reward}
    r(i,e_i,d_i) = \begin{cases} 
    - procT, & \text{if } e_i=0 \land d_i=0\\
    + procT, &  \text{if } e_i=0 \land d_i=1\\
    - retT(i, procT), & \text{if } e_i=1 \land d_i=1\\
    + retT(i, procT), & \text{if } e_i=1 \land d_i=0\\
\end{cases}
\end{equation}

The training of the agent occurs in cycles, called episodes.
Each  
episode is  
completed  
when the agent has been trained for all  
points of interest using as input the experience gained during the last mission. The agent is trained for multiple episodes, depending on the desired number of training timesteps, i.e.,  
prediction attempts  
made during (re)training to receive the respective rewards and learn to take better decisions.

With respect to  
Algorithm~\ref{alg:control}, the (re)training process 
takes place within the $\Call{MissionEnd}$ method of the \textsc{DecisionMethod} component, at the end of the mission. In other words, when using the RL decision method, the training is performed offline when the drone has landed (and maybe re-charging its batteries),  
thus even heavyweight and time-consuming training 
will not affect the current or the next mission. Also, in our case, the agent is trained incrementally. 
More specifically, after each mission, the agent is re-trained 
merely using 
the experience of the last mission, rather than training the RL agent each time from scratch using all previous (and ever increasing) historical data. This induces minimal overhead 
without affecting the life cycle of drone operation. The retrained agent is loaded before the next mission starts, via the $\Call{MissionBegin}$ method, while the prediction at each point of interest is made by invoking the model via the $\Call{Decide}$ method. Finally, the $\Call{Feedback}$ method records the outcome so that this information is used in the re-training when the mission is completed.

\subsection{Algorithms}
\label{subsec:rl_algos}

We explore RL in conjunction with $3$ popular model-free RL algorithms: DQN~\cite{mnih2013playing},
A2C~\cite{mnih2016asynchronous} and PPO~\cite{schulman2017proximal}. 
These are implemented using the Stable Baselines3 library~\cite{StableBaselines3}. 

\textbf{Deep Q-Network (DQN)} is an off-policy RL algorithm that  
manages discrete action spaces. It leverages two neural networks: (i) a Q-network that estimates the action-value function Q, which maps state/action pairs to their expected discounted cumulative long-term reward, and (ii) a target network that is  
updated periodically but less frequently compared to the Q-network. In addition,  
it uses a replay buffer to store previous interactions with the environment so that they can be reused during training.
The above characteristics render DQN sample-efficient. 
Replay memory also de-correlates previous decisions from each other by sampling them randomly in the training process. 

\textbf{Advantage Actor-Critic (A2C)} is an actor-critic method that can handle discrete or continuous action spaces.  
It is a synchronous version of the Asynchronous Advantage Actor-critic (A3C) method, combining the advantages of both value- and policy-based methods. More specifically, it exploits the actor, a component that uses the current policy to decide which action to perform, while the critic evaluates the actions taken by the actor and provides relevant feedback.

\textbf{Proximal Policy Optimization (PPO)} is  
based on stochastic gradient ascent 
optimizing a surrogate objective function. Like A2C, it can manage discrete or continuous action spaces.  
It is also more robust  
(less prone to instability) mainly  
thanks to its conservative policy updates, which are not allowed to diverge more than a certain threshold by applying a clipping function. 

\section{Evaluation}
\label{sec:eval}

\begin{figure}
  \begin{subfigure}[c]{.4\linewidth}
    \centering
    \includegraphics[width=\linewidth]{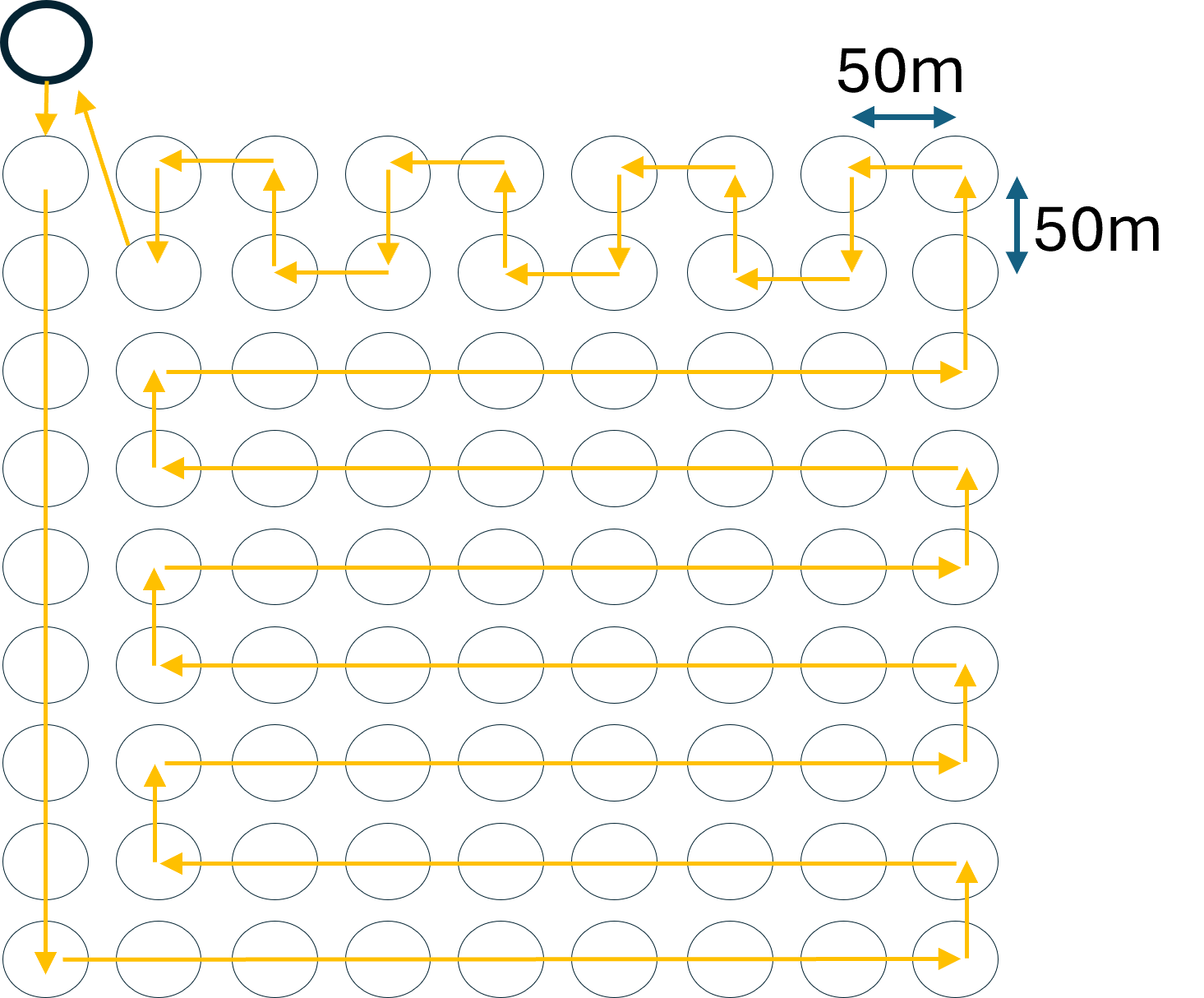}%
    \caption
      {%
        Points of interest and mission path (same for all missions).
        \label{fig:grid}%
      }%
  \end{subfigure}\hfill
  \begin{tabular}[c]{@{}c@{}}
    \begin{subfigure}[c]{.55\linewidth}
      \centering
      \includegraphics[width=\linewidth]{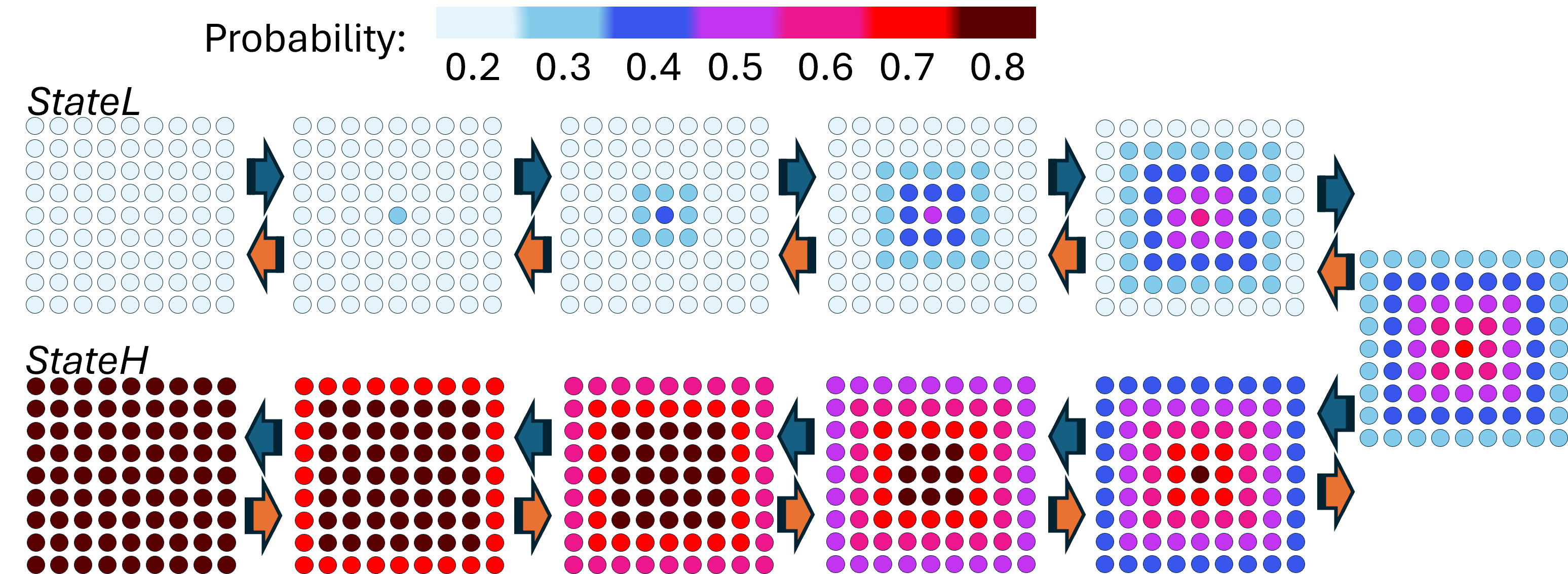}%
      \caption
        {%
          Change of event detection probability at the points of interest according to pattern A.
          \label{fig:patA}%
        }%
    \end{subfigure}\\
    \noalign{\bigskip}%
    \begin{subfigure}[c]{.55\linewidth}
      \centering
      \includegraphics[width=\linewidth,page=2]{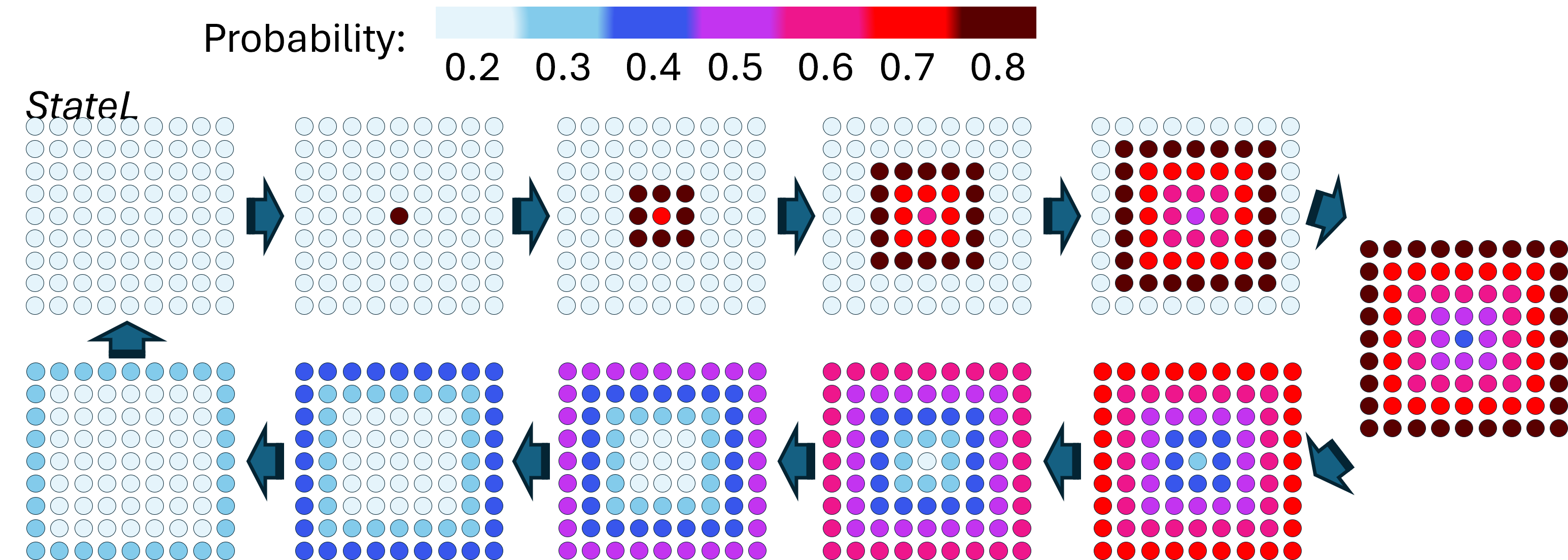}%
      \caption
        {%
          Change of event detection probability at the points of interest according to pattern B.
          \label{fig:patB}%
        }%
    \end{subfigure}
  \end{tabular}
  \caption
    {%
      Mission area and % 
      patterns of event detection probability change. %
      \label{fig:expsetup}%
    }
\end{figure}

For our evaluation, we use the same simulator as in~\cite{staygo}, with realistic flight-related delays 
$flyT_{i,j}$ and $retT(i,procT)$, obtained from experiments with a real quadcopter drone in the field, configured to fly at a cruising speed of $4$m/s. We also use the same settings for the rest of the delays, $senseT=1$~sec, $procT=10$~sec, and $actT=10$~sec. Next, we present the experimental setup and discuss the results achieved by the proposed machine-learning approaches vs other benchmarks. 

\subsection{Topology and change of event probability}
\label{sec:expsetup}

The target area consists of 81 points of interest, arranged in a $9\times9$ grid as shown in 
Figure~\ref{fig:grid}. 
The vertical and horizontal distance between points is 50 meters. The home point, from which the drone takes-off and where it returns to land, is the bold node at the periphery of the area of interest. The yellow arrows indicate the drone's path in all missions.

We assume that the occurrence of events at each point of interest is governed by an underlying probability. Furthermore, we let this probability change between missions according to two different patterns, shown in Figure~\ref{fig:patA} and Figure~\ref{fig:patB}. Both start from and end at StateL where all points have low event probability $0.2$. In pattern~A, the probabilities increase  
gradually  
from the center to  
the periphery of the mission area, eventually leading to StateH where all points have a high probability $0.8$. 
Then, the probabilities decrease gradually in the reverse direction, eventually going back to StateL. In pattern~B, probabilities also increase 
in the same direction.
However,  
unlike pattern~A, this increase 
occurs  
abruptly,  
from low ($0.2$) to high ($0.8$).   
Also, the probabilities decrease gradually from the center to the periphery, 
in the opposite direction than in pattern~A.

Further, we experiment with  
two rates of probability change.  
In the \textit{fast} rate, 
the  
probability at each point of interest changes between missions in steps of $0.1$, as shown in Figure~\ref{fig:patA} and Figure~\ref{fig:patB}. In the \textit{slow} rate, the probability changes each time by $0.05$. As a result, for each transition shown in the figures between a state where a given point of interest has probability $p_1$ and a state where the same point has probability $p_2$, there is an intermediate state where this point has probability $(p_1 + p_2) / 2$. These intermediate states are not shown in the figures for brevity. Note that scenarios with the slow rate include twice the number of states (and missions) vs the fast rate. 

In our experiments, we assume that the drone  
executes the same mission plan periodically, visiting each time the same points of interest in the same pre-specified order. The goal of our evaluation is to see how well each decision method performs in each mission for the respective state (the event probability at each point of interest) and how well it adapts to more or less abrupt changes (in these probabilities).

\begin{table}
\centering
\begin{tabular}{|l|c|c|c|}
\hline
\textbf{Parameter} & \textbf{DQN} & \textbf{A2C} & \textbf{PPO} \\ 
\hline
 Neural Network & MLP & MLP & MLP\\
\hline
 Number of hidden layers & $2$ & $2$ & $2$ \\
\hline
 Units (per hidden layer) & $64$ & $64$ & $64$ \\
\hline
 Optimizer & Adam & RMSprop & Adam \\
 \hline
 RMSprop $\epsilon$ & N/A & %$1e-5$ 
 $0.00001$ & N/A \\
\hline
 Activation function & ReLU & tanh & tanh \\
 \hline
 Target network update interval & $10000$ & N/A & N/A \\
\hline
 Learning rate  & $0.0001$ & $0.0007$ & $0.00005$ \\
\hline
 Discount factor & $0.99$ & $0.99$ & $0.99$ \\
 \hline
 Replay buffer size  & $1000000$ & N/A & N/A \\
 \hline
 Minibatch size & $32$ & N/A & $64$ \\
 \hline
 Soft update coefficient  & $1.0$ & N/A & N/A \\
 \hline
 Exploration fraction  & $0.1$ & N/A & N/A \\
 \hline
 Initial value of random action probability & $1.0$ & N/A & N/A \\
 \hline
 Final value of random action probability & $0.05$ & N/A & N/A \\
 \hline
 Generalized advantage estimator Lambda & N/A & $1.0$ & $0.95$ \\
 \hline
 Entropy coefficient for loss calculation & N/A & $0.0$ & $0.0$ \\
\hline
 Value function coefficient for loss calculation & N/A & $0.5$ & $0.5$ \\
 \hline
 Number of epochs when optimizing the surrogate loss & N/A & N/A & $10$ \\
 \hline
 Max value for gradient clipping & $10$ & $0.5$ & $0.5$ \\
 \hline
Number of timesteps (per mission) & $100000$ & $100000$ & $10000$ \\
\hline
\end{tabular}
%}
\caption{RL agent learning parameters for each algorithm.}
\label{tab:drl_parameters}
\end{table}

\subsection{Fine-tuning}

The  
proposed methods  
were  
fine-tuned  
after extensive experimentation  
(not included for the sake of brevity). The respective configuration parameters are briefly presented below.

\textbf{Perceptron.}
We set the bias  
$B=0.001$,  
forcing the predictor to initially take \textit{stay} decisions. The neighborhood boundary  
is set to $D=150$~meters  
and we use the $H=2$ most recent outcomes at each point of interest. 
Finally, for the weight training we set the learning rate to $\eta=0.1$. 

\textbf{Reinforcement learning.}
The hyperparameter settings that were selected after fine-tuning for the three RL-based methods 
are shown in Table~\ref{tab:drl_parameters}.  
The selection  
criterion is to minimize the number of opposite decisions  
and the  
increase in mission time 
vs the knowledgeable method (discussed in Section~\ref{sec:benchmarks}). The chosen number of timesteps (per mission) is sufficient to  
achieve proper training and generalization of the models (we have verified that increasing the number of timesteps does not improve performance). Notably, the best version of PPO requires 10x fewer timesteps than DQN and A2C.

\subsection{Benchmarks}
\label{sec:benchmarks}

Besides the proposed ML approaches, we also use the following methods as benchmarks for comparison. 

\textbf{2-bit Predictor:} This method uses a 2-bit predictor~\cite{lee1984branch} for each point of interest. The predictor consists of a saturating counter with states $\{00,01,10,11\}$ (in binary), with the most significant bit  
giving the prediction. 
When the drone does not detect an event,  
the counter is increased and the state changes towards the states that predict $1$ (go). Conversely, when the drone detects an event that must be handled, the counter is decreased and the state changes in the reverse direction towards the states that predict $0$ (stay). For the first mission,  
the counter is initialized to $01$, forcing the drone to wait. 

\textbf{Regression:}  
Proposed in~\cite{staygo},  
this method predicts the event probabilities at each point  
by running a regression algorithm on the experience gained from previous missions.  
The decision  
is taken by weighing the predicted probability  
with the time penalties for the drone if the decision is wrong.  
Also, an anomaly detection mechanism is used to detect whether the previous experience is invalid,  
in which case the memory of the algorithm is reset. For the first mission, 
the method is initialized to choose to wait (stay) at all points. 

\textbf{Knowledgeable:}  
This method  
takes decisions like the above  
approach, but has perfect knowledge about the underlying event probability at each point of interest (there is no need to learn/infer this by observing the presence or absence of events at runtime). Given that in the general case it is not possible to have such perfect knowledge in reality, we use this as an \textit{idealized} reference for all other methods. Note, however, that 
this method is \textit{not} an oracle, thus it can still take wrong decisions  
(more likely when the event probability is  
near $0.5$).

\begin{figure*}
    \centering
    \begin{subfigure}{0.95\linewidth}
\includegraphics[width=\linewidth]{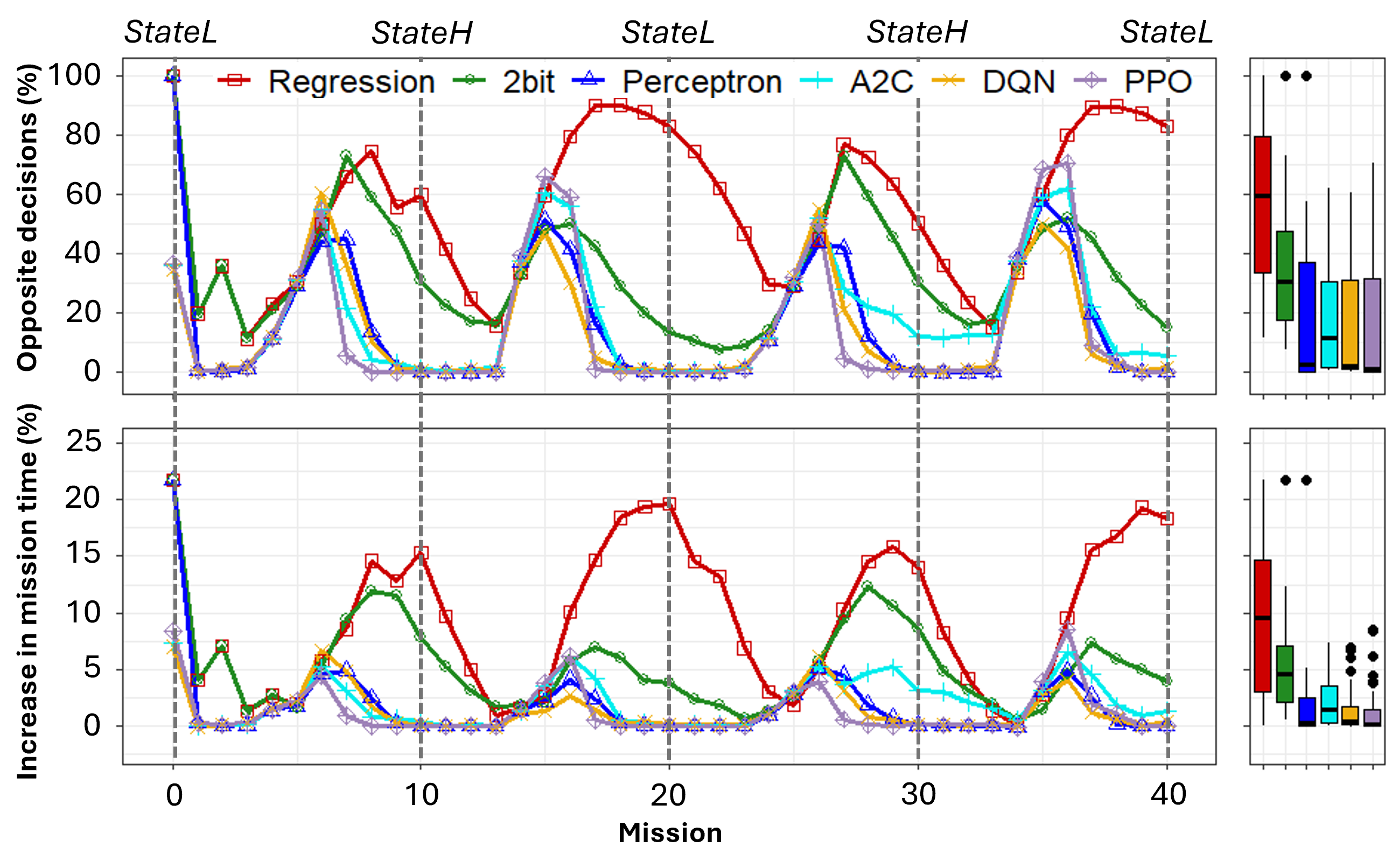}
\caption{Fast rate of event probability change}
\label{fig:senarioArate01}
\end{subfigure}
\begin{subfigure}{0.95\linewidth}
\includegraphics[width=\linewidth]{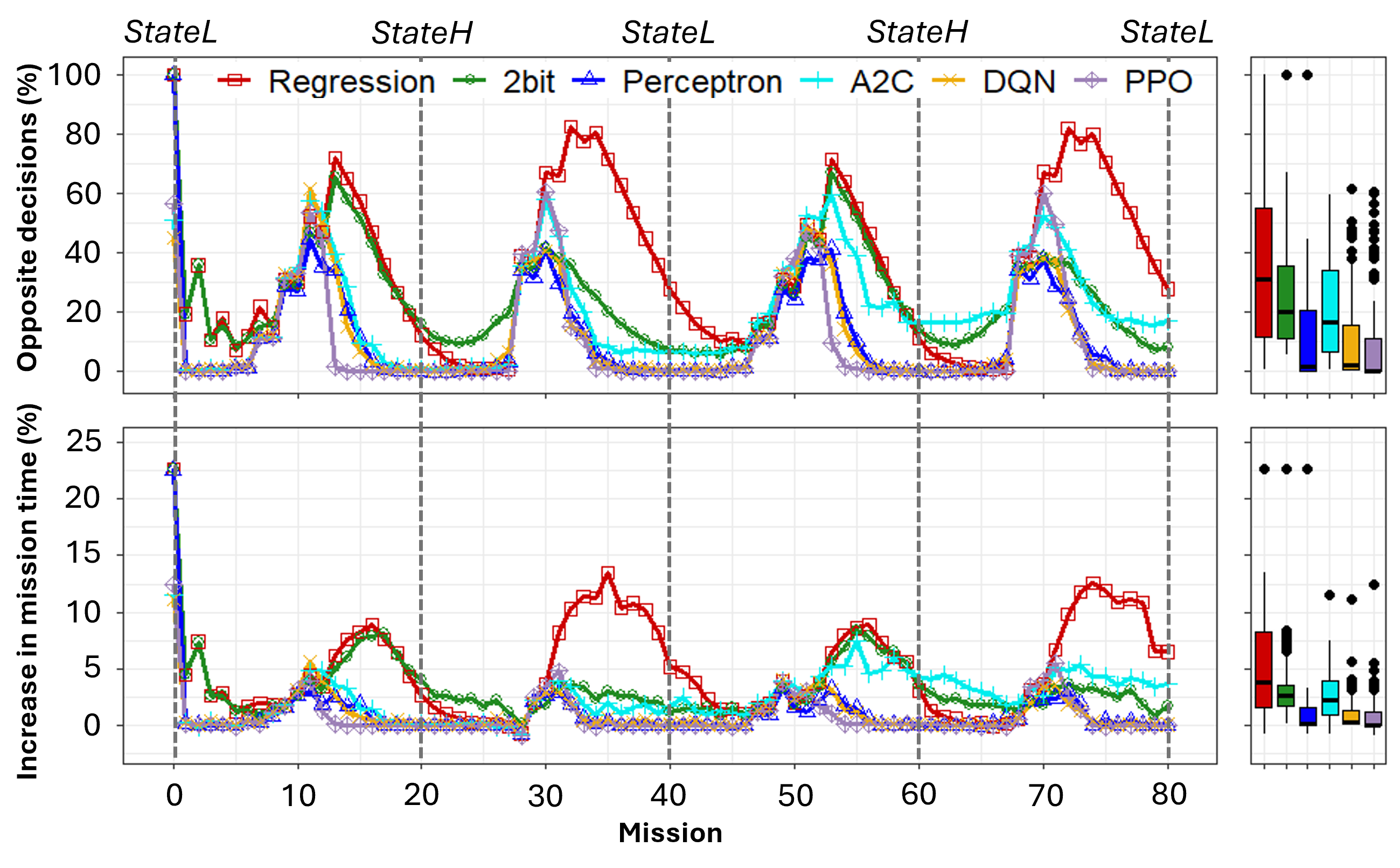}
\caption{Slow rate of event probability change}
\label{fig:senarioArate005}
\end{subfigure}
    \caption{Pattern A: opposite decisions (top) and mission time increase (bottom)  
    vs the knowledgeable method.}
    \label{fig:scenarioA}
\end{figure*}

\begin{figure*}
    \centering
    \begin{subfigure}{0.95\linewidth}
\includegraphics[width=\linewidth]{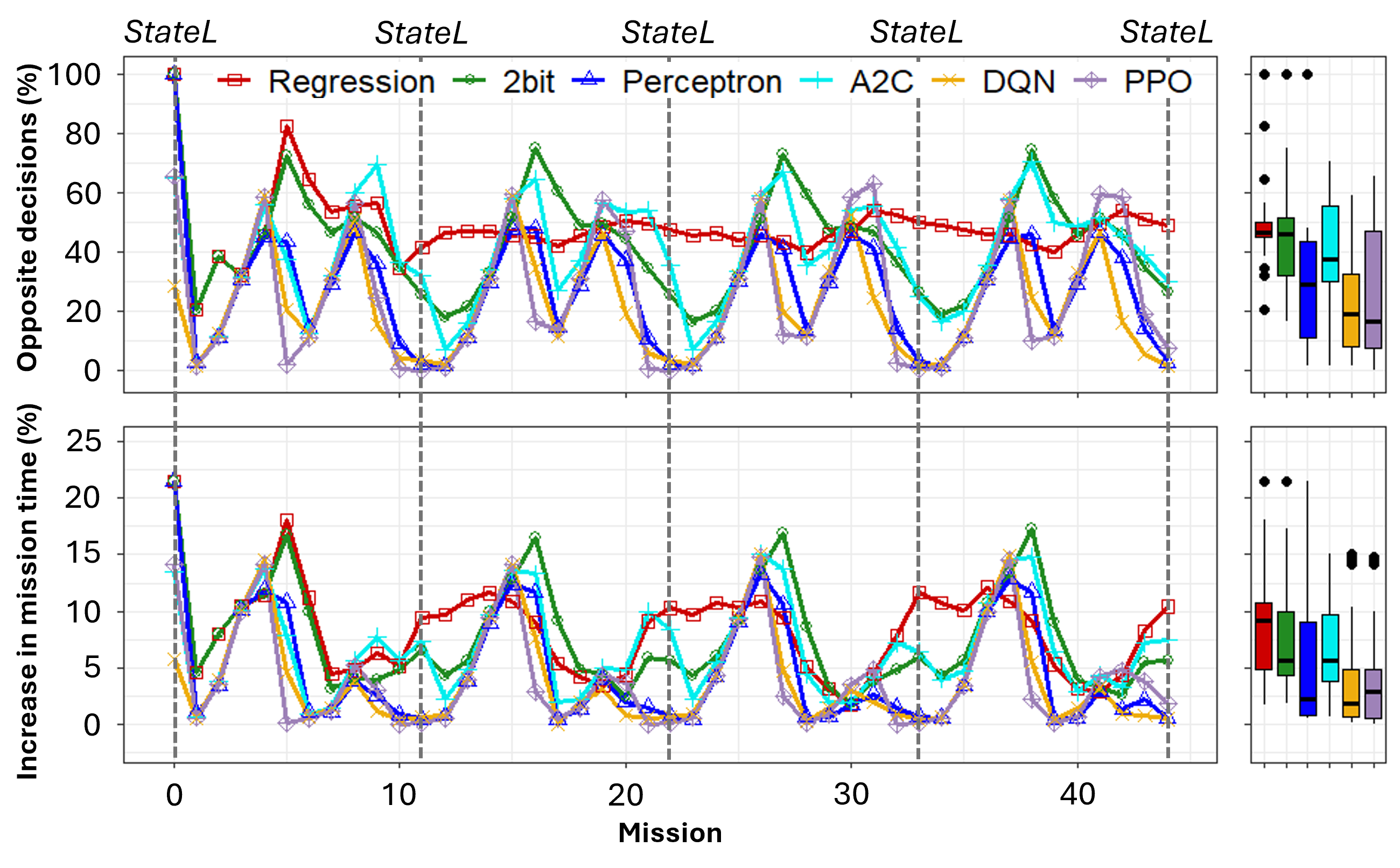}
\caption{Fast rate of event probability change}
\label{fig:scenarioBrate01}
\end{subfigure}
\begin{subfigure}{0.95\linewidth}
\includegraphics[width=\linewidth]{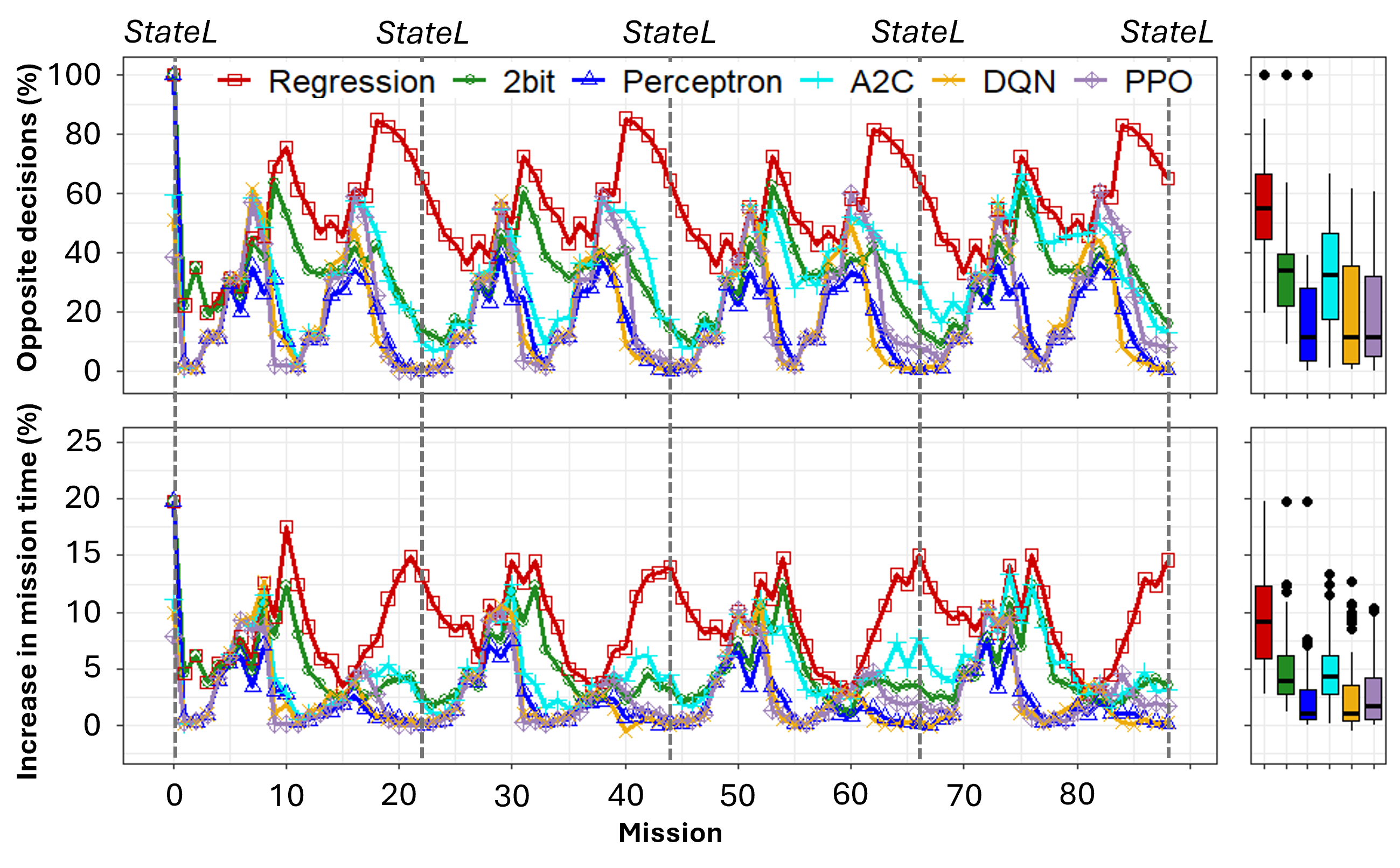}
\caption{Slow rate of event probability change}
\label{fig:senarioBrate005}
\end{subfigure}
    \caption{Pattern B: opposite decisions (top) and mission time increase (bottom)  
    vs the knowledgeable method.}
    \label{fig:scenarioB}
\end{figure*}

\subsection{Results}

We run every  
scenario  
for each probability change rate $20$ times, for a different randomly chosen event  
occurrence at each point (generated based on the respective underlying probability at each state), and report the average over the $20$ runs. To ensure a fair comparison, the occurrence of an event at each point is generated offline and is stored in a file, which is replayed when testing each decision method.

The results are shown in Figure~\ref{fig:scenarioA} for pattern~A and in Figure~\ref{fig:scenarioB} for pattern~B. In each figure, the top plots show the opposite decisions vs the knowledgeable method in each mission  
(probability state) as described in Section~\ref{sec:expsetup}. Note that fewer opposite decisions correspond to better performance (shorter mission times) as the knowledgeable method takes informed decisions based on perfect knowledge about the underlying event probabilities. 
The plots at the bottom show the relative increase in the mission time vs the knowledgeable method for each mission (again, lower is better). 
Note that the experiments for pattern~A run for two StateL $\rightarrow$ 
StateL cycles, while pattern~B experiments run for four  
cycles. We accompany each line plot with a boxplot to visualize the statistical performance  
over all mission executions.
In addition, as key performance metrics, we report the maximum (worst-case) and median increase in the mission time vs the knowledgeable method, summarized in Table~\ref{tab:results} for each scenario  
(we ignore the first mission where there is no prior experience and methods take a default decision at all points, predetermined by their bias). Next, we discuss the results for each pattern separately, followed by an overall assessment.

\begin{table}[htbp]
\centering
\begin{tabular}{|l|c|c|c|c|c|c|c|}
\hline
 & & \textbf{Regression} & \textbf{2bit} & \textbf{Perceptron} & \textbf{A2C} & \textbf{DQN} & \textbf{PPO} \\
\hline
Pattern A fast rate 
& \makecell{max \\ median} 
& \makecell{19.6\% \\ 9.1\%} 
& \makecell{12.3\% \\ 4.4\%} 
& \makecell{\textbf{5.1\%} \\ 0.3\%} 
& \makecell{6.5\% \\ 1.4\%} 
& \makecell{6.7\% \\ 0.4\%} 
& \makecell{8.5\% \\ \textbf{0.1\%}} \\
\hline
Pattern A slow rate 
& \makecell{max \\ median} 
& \makecell{13.5\% \\ 3.8\%} 
& \makecell{8.4\% \\ 2.6\%} 
& \makecell{\textbf{3.3\%} \\ 0.1\%} 
& \makecell{7.5\% \\ 2.2\%} 
& \makecell{5.6\% \\ 0.3\%} 
& \makecell{5.5\% \\ \textbf{0.0\%}} \\
\hline
Pattern B fast rate 
& \makecell{max \\ median} 
& \makecell{18.1\% \\ 9.2\%} 
& \makecell{17.3\% \\ 5.7\%} 
& \makecell{\textbf{13.3\%} \\ 2.1\%} 
& \makecell{15.1\% \\ 5.6\%} 
& \makecell{15.1\% \\ \textbf{1.7\%}} 
& \makecell{14.8\% \\ 2.7\%} \\
\hline
Pattern B slow rate 
& \makecell{max \\ median} 
& \makecell{17.5\% \\ 9.2\%} 
& \makecell{12.4\% \\ 3.9\%} 
& \makecell{\textbf{7.6\%} \\ \textbf{1.1\%}} 
& \makecell{13.4\% \\ 4.3\%} 
& \makecell{12.7\% \\ \textbf{1.1\%}} 
& \makecell{10.3\% \\ 1.7\%} \\
\hline
\end{tabular}
%}
\caption{Max/median relative increase in mission time vs the knowledgeable method (best values in bold).}
\label{tab:results}
\end{table}

\subsubsection{Pattern A (Figure~\ref{fig:scenarioA})}

We observe the same high-level  
trend for all methods. In each StateL $\rightarrow$  
StateL cycle, there 
are two phases where performance drops and then improves again. This is because the gradual changes in probability eventually flip the decision that must be taken at each point, from go to stay and vice versa. Also, each such phase includes several states where many points have probability equal or close to $0.5$, making it harder to predict this correctly and take the same decisions as the knowledgeable method. Note that these peaks of opposite decision making  
lead to corresponding increased mission times, but to a smaller degree. The reason is two-fold. On the one hand, as previously noted, the knowledgeable method can take wrong decisions (and this  
becomes more likely  
when the event probability is around $0.5$). On the other hand, the mission time  
includes delays that can not be  
avoided (like $flyT_{i,j}$, $senseT$ or $actT$) thereby decreasing the negative impact of wrong decisions. 
The same  
trend holds for the slow rate, but with improved decision performance and reduced mission times, as changes are smoother and there are more opportunities to learn. 

Clearly, the regression 
cannot handle changes well, especially during the aforementioned phases, 
thus producing noisy estimations about the  
event probability  
at each point. In turn, this results in uninformed decisions and increased mission times. The 2-bit predictor performs consistently better, 
but is still suboptimal in almost every mission. The perceptron is significantly better than the previous methods, especially in the challenging state transitions where it adapts faster, achieving a median mission time very close to the knowledgeable method. 

Looking at the RL algorithms,  
one can observe that both DQN and PPO perform well, 
achieving very good performance in terms of the median mission time that is close (DQN) or even better (PPO) than the perceptron, albeit having  
higher worst-case mission times. 
A2C initially performs close to the other RL algorithms during the first StateL $\rightarrow$ 
StateL cycle, but  
shows a noticeable performance drop in the second cycle.

\subsubsection{Pattern B (Figure~\ref{fig:scenarioB})} 

The high-level  
trend in each StateL $\rightarrow$ StateL cycle is 
similar to that of pattern~A. Namely, in each such cycle, there are two major transition phases, in the first phase from states where in most points the right decision is to go towards states where the right decision is to stay, and vice versa in the second phase. However,  
the probabilities change more abruptly  
during the first phase 
while the entire StateL $\rightarrow$ StateL cycle has half the transitions compared to pattern~A.
This makes learning harder, resulting in worse decision performance and increased mission times compared to pattern~A.  
The two peaks of bad 
performance in each cycle, corresponding to the two aforementioned phases, 
are clearly visible in the plots. Note that taking opposite decisions vs the knowledgeable method has a higher impact on the mission time increase in the first phase compared to the second phase of each cycle. This is because the time penalty of a wrong go decision (in the first phase, the drone has learned to go, but due to the change in probabilities the best decision is to stay)
is higher than that of a wrong stay decision (in the second phase, the drone has learned to stay, but the probability change makes go the best decision). 

Under these more challenging circumstances, all methods exhibit a worse performance than in pattern~A 
(as a minor exception, in the  
fast rate, the regression has a slightly better worst-case mission time). Like in the previous experiments, the regression does not manage to adapt 
well, 
taking many opposite decisions and resulting in significantly increased mission time vs the knowledgeable method. The 2bit and A2C achieve better results than the regression, but still have relatively poor performance, especially in terms of the median increase of the mission time. Once again, the 
perceptron, DQN and PPO consistently  
outperform all other methods, with results quite close to those of the knowledgeable method. 

\subsubsection{Overall}

The perceptron, DQN, and PPO 
achieve the best results over all 
scenarios, with a big difference compared to the regression. On the one hand, the  
worst-case mission time in the state transitions where the probabilities and the right decisions are harder to predict, is significantly lower. 
More precisely, 
the  
improvement vs the regression across all scenarios and rates is  
$1.4x$ up to $4.1x$ for the perceptron, $1.2x$ up to $2.9x$ for DQN, and $1.2x$ up to $2.5x$ for PPO. 
On the other hand, these methods manage to overcome the periods of bad 
predictions and converge fast to good 
decisions, which, in turn, translates to reduced mission times. 
This can be  
confirmed by looking the median increase in the mission time vs the knowledgeable method. Namely, in pattern~A, all three methods  
achieve practically the same median mission times as the knowledgeable method.
In pattern~B, 
despite having a more notable mission time increase vs the knowledgeable method, the perceptron, DQN and PPO respectively outperform the regression by $4.4x$, $5.4x$ and $3.4x$ for the fast rate, and by $8.4x$, $8.4x$ and $5.4x$ for the slow rate. Note that the improvement  
is significantly higher for the slow rate, showing that when there is sufficient opportunity to learn, these methods manage to do this much better than the regression. 

Comparing the perceptron with the two RL methods, 
DQN and PPO  
achieve a slightly better median increase in mission time for the fast rate of event probability change, in pattern~B and pattern~A, respectively. For the slow rate, the perceptron has 
very similar  
performance. 
Further, the perceptron consistently has the lowest 
worst-case increase in mission time over all scenarios and rates. 
We conclude that the perceptron 
is more stable 
and robust to abrupt changes, while DQN/PPO seem  
to adapt faster to complex and highly dynamic environments. Between the two RL algorithms, PPO is more attractive due to the much faster re-training time ($10x$ fewer timesteps).

\section{Conclusion}
\label{sec:concl}

Focusing on data-driven drone missions, we have proposed different machine-learning methods for  
deciding whether to wait for data processing to finish before moving to the next point of interest, or to start this movement while the computation is still being performed in anticipation that processing will not reveal an event or situation that requires extra handling at the previous point of interest. Our evaluation for scenarios where the probability of  
such event  
occurrence at each point of interest changes with time, show that these methods can  
achieve significantly better results than the regression-based method previously proposed in the literature, while  
performing close to a method that has perfect knowledge of the underlying event probabilities. 

Regarding future work, in order to further evaluate each decision method, we plan to experiment with a wider variety of scenarios, where the event probabilities follow different and even completely random patterns. Another interesting direction is to support more decisions and control options. For instance, the drone could set a lower cruising speed when deciding to go but the probability of an event being detected at the previous point of interest is non-negligible, thereby reducing the time penalty in case the drone must indeed return to the previous point of interest to take further action.

\section*{Acknowledgments}

This work has been supported by the Horizon Europe research and innovation programme of the European Union, project MLSysOps, grant agreement number 101092912.

\bibliographystyle{splncs04}
\bibliography{bib}

\end{document}